\documentclass[sigconf, screen]{acmart}
\AtBeginDocument{%
  \providecommand\BibTeX{{%
    \normalfont B\kern-0.5em{\scshape i\kern-0.25em b}\kern-0.8em\TeX}}}

\usepackage{tikz}

\setcopyright{none}
\renewcommand\footnotetextcopyrightpermission[1]{}
\acmConference[]{}{}{}
\begin{document}

\title{TropeTwist: \\ Trope-based Narrative Structure Generation}
\author{Alberto Alvarez}

\affiliation{%
  \institution{Malmö University, Game Lab}
  \city{Malmö}
  \country{Sweden}
  \postcode{21119}
}
\email{alberto.alvarez@mau.se}

\author{Jose Font}
\affiliation{%
  \institution{Malmö University, Game Lab}
  \city{Malmö}
  \country{Sweden}
  \postcode{21119}
}
\email{jose.font@mau.se}

\renewcommand{\shortauthors}{Alvarez and Font}

\begin{abstract}
Games are complex, multi-faceted systems that share common elements and underlying narratives, such as the conflict between a hero and a big bad enemy or pursuing a goal that requires overcoming challenges. However, identifying and describing these elements together is non-trivial as they might differ in certain properties and how players might encounter the narratives. Likewise, generating narratives also pose difficulties when encoding, interpreting, and evaluating them. To address this, we present TropeTwist, a trope-based system that can describe narrative structures in games in a more abstract and generic level, allowing the definition of games' narrative structures and their generation using interconnected tropes, called narrative graphs. To demonstrate the system, we represent the narrative structure of three different games. We use MAP-Elites to generate and evaluate novel quality-diverse narrative graphs encoded as graph grammars, using these three hand-made narrative structures as targets. Both hand-made and generated narrative graphs are evaluated based on their coherence and interestingness, which are improved through evolution.
\end{abstract}




\maketitle

\section{Introduction}

\begin{figure*}
    \centering
    \includegraphics[width=0.95\textwidth]{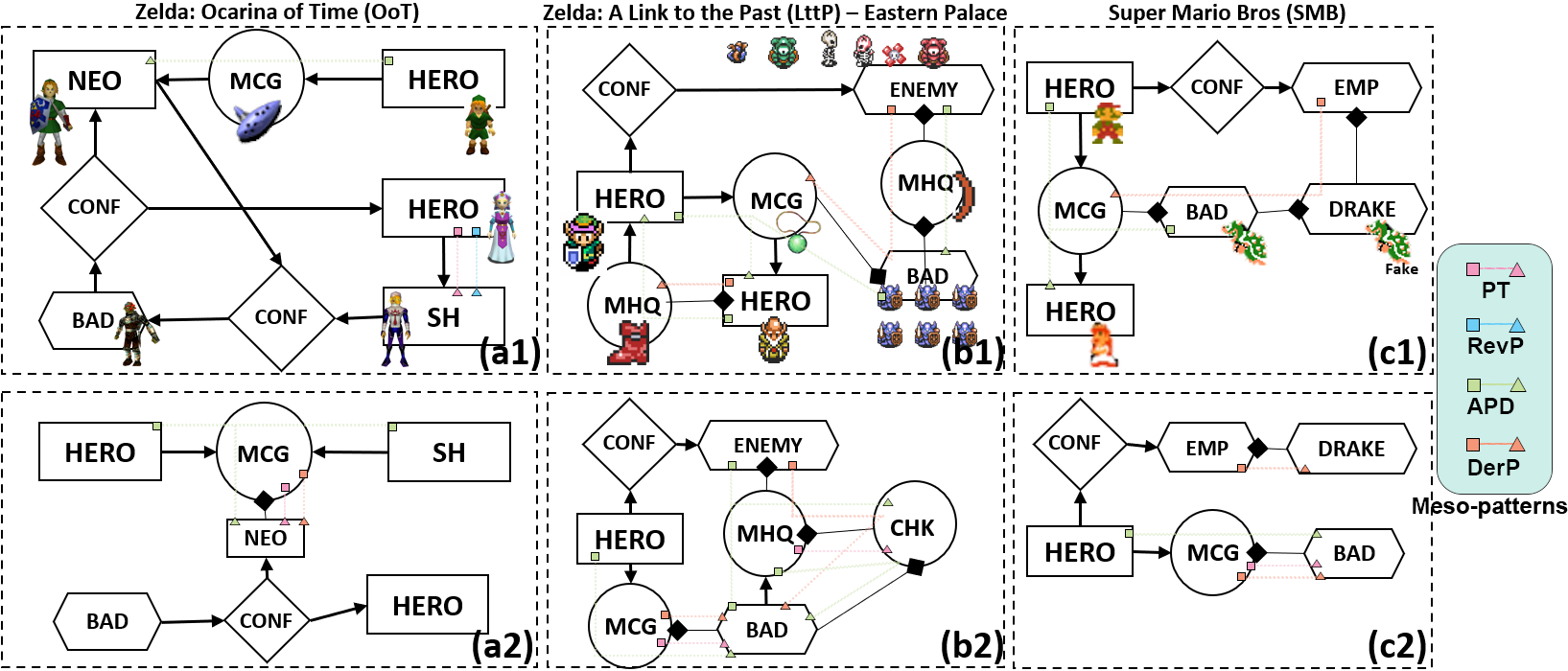}
       \caption{Proof-of-concept narrative structures of existing games (top row) created using TropeTwist with the available nodes (table~\ref{tab:tropes}). Bottom row shows exemplar elites generated with MAP-Elites using the respective top row narrative structure as root. Color matching squares, lines, and triangles denote different meso-patterns in the structures. Squares and triangles are the start and end of a meso-pattern, respectively.}
       \label{fig:teaserfig}
\end{figure*}

There exists a plethora of games\footnote{For instance, currently there are more than 68k games in steam \url{https://store.steampowered.com/search/?category1=998}.}, with diverse genres and each containing a different set of gameplay mechanics, audio, level, graphic, and narrative facets. The creation and combination of these facets make game development a hard task, commonly involving a diverse group of developers~\cite{Blow2004-gamesHard}. Likewise, the generation of these facets in conjunction has been categorized as one of the biggest and most challenging tasks within computational creativity~\cite{Liapis2014-gameCreativity,Liapis2019-OrchestratingGames}. However, games share common elements and underlying narratives, but it is non-trivial how to identify these, how to define and analyze these games structurally, or what type of common underlying structures exist; pointed out as well by~\cite{aarseth2018-ontologicalMeta,vozaru_game_2022}.

Among the different facets, narrative stands out in games as it helps to create meaning, make sense of situations, and make games [stories] recognizable~\cite{mateas2003-facade,Aarseth2012-Narrativetheory,kybartas2016survey,flodtol2020-WIPMakeSenseDungs}. Narrative structures can be used to describe how an experience or story is to be developed as argued by Barthes~\cite{Barthes75-introStructNarr}, and to create an abstract representation based on the narrative structure instead of a temporal and partially-ordered sequence of events~\cite{Szilas2003-structuralModelsIDtension}. Common narrative structures used in many domains are Aristotle's drama structure, which subdivides a story into \textit{exposition}, \textit{climax}, and \textit{resolution} or Propp's analysis on the morphology of russian folktale, which revealed a common structure among them, denoted as Propp's 31 ``narremes''~\cite{propp1975-morphology}.

This paper presents \emph{TropeTwist}, a preliminar system that uses Tropes~\cite{Thompson2018-usingTropesNarrativeEvents,tropesSimpsons} extracted from TvTropes~\cite{tvtropes,periodicTable} as patterns and fundamental units, which when combined can compose structures further representing other composed tropes. Common narrative structures can be identified and defined using \emph{TropeTwist}. TropeTwist can define generic aspects of a story, leading to the identification of events, roles, and narrative elements, as well as a novel way to form narratives. As a proof-of-concept, we built, analyzed, and described structurally three game examples shown in figure~\ref{fig:teaserfig}, top row.


We propose graph grammars as indirect encoding of narrative graphs and the use of the Multi-dimensional Archive of Phenotypic Elites (MAP-Elites)~\cite{Mouret2015-MAPElites} to generate novel variations (shown in figure~\ref{fig:teaserfig}, bottom row) using the proof-of-concept examples as roots. Simultaneously, we propose metrics to evaluate the resulting narrative graphs' coherence, cohesion, and interestingness. Our preliminary results show that we can produce more interesting structures retaining coherence based on our metrics. 

\section{Related Work}



Propp~\cite{propp1975-morphology} analyzed Russian folktales identifying their fundamental structure in 31 steps. His work contributed to the identification of core elements, the proposal of actions and events as \emph{functions} and narrative atoms, and roles that are recurrent within the folktales. Propp emphasized that these 31~\emph{functions} and their arrangement were the structure and what gave meaning to the story discourse. Barthes~\cite{Barthes75-introStructNarr} proposed three intertwined and progressively integrated levels in narrative work: \emph{functions}, \emph{actions}, and \emph{narration}. His work is characterized by the proposal of fundamental narrative units in the~\emph{function} level to better assess and identify structures in a narrative. Furthermore, Baikadi and Cardona-Rivera~\cite{Baikadi2012-Narreme} further discuss these fundamental units as~\emph{narremes} encoding narrative state and how they could be combined to narrative structures. Their work, similar to TropeTwist, proposes a graph structure of interconnected~\emph{narremes}. However, they defined narrative axes like Barthes, where each connection between~\emph{narremes} means a change along a narrative axis. In games, the narrative is usually directed by quests, which Aarseth~\cite{aarseth2005hunt} discusses as a central element in games to make sense of other elements, and which are defined by Yu et al. as a form of structure, dividing the story into achievable rewards and partially ordered set of tasks~\cite{yu2020quest}.

Furthermore, the generation of narratives, stories, and quests using a variety of techniques such as planning algorithms~\cite{Riedl2006-StoryPlanningCreativity,young2013-plansNarrGen}, grammars~\cite{hartsook2011-storyWorlds,Alvarez2021-questgram}, or machine learning~\cite{tambwekar2019-controllableNeuralStory,vanstegeren2021-gpt2quests}, is a growing and important field within games research and narrative research in general~\cite{Gervas2009-ComputationalStoryCreativity,kybartas2016survey,yu2020quest,Eladhari2014-storymakinggames}. One typical approach for the generation of content and stories is the use of patterns representing different elements such as level design patterns~\cite{alvarez2019empowering,flodtol2020-WIPMakeSenseDungs}, quest patterns and common quests in games~\cite{Trenton2010-questpatterns,Doran2011-questsMMORPGs}, or identifying fundamental units and assembling them based on various pre-conditions~\cite{Kreminski2018-SketchingStorylets,Garbe2019-StoryletsAssembler}. A particular type of pattern is tropes, which are concepts that are recurrently used in transmedia storytelling~\cite{tropesSimpsons,tvtropes}. Horswill~\cite{Horswill2016-DearLeaderTrope} focused on constructing an expressive language that could encode plot tropes as story fragments, composing a database of fragments combined sequentially with a planner. Similarly, Thompson et al.~\cite{Thompson2018-usingTropesNarrativeEvents} used the idea of tropes as story bits where a system would construct valid stories from users' defined story bits with pre-and post-conditions. TropeTwist uses the idea of tropes for nodes and patterns in structures and encodes and represents these as a graph. \emph{Scheherazade} is a system that can capture narrative structures by encoding and annotating narrative texts, which introduced the Story Intention Graph model, a formal and expressive representation of narratives~\cite{elson-2012-dramabank}.

Moreover, we use graph grammars and grammar recipes to generate structures. This approach is similar to how Dormans and Bakkes \cite{dormans2011generating} generate missions and space using a ``key and lock'' structural idea. Our approach uses MAP-Elites, a quality-diversity algorithm that uses behavioral dimensions that are orthogonal to the objective function to store diverse individuals in a grid~\cite{Mouret2015-MAPElites}. Evolutionary algorithms are a popular approach in PCG to generate diverse type of content~\cite{Togelius2011}, but not as much for narrative content. MAP-Elites have been used to generate content in different game facets such as levels~\cite{charity2020baba,Alvarez2020-ICMAPE}, mechanics~\cite{charity2020mech}, or enemy behavior~\cite{Khalifa2018}.



Assessing narratives is a complex and non-trivial task. The goal is to create a narrative that is both syntactically correct (e.g., coherent and consistent) and semantically rich (e.g., novel and interesting)~\cite{Rowe2009-STORYEVAL,Hargood2011-NarrativeCohesion,Castricato2021-FabulaStoryCoherenceMeasure}. Perez y Perez and Ortiz~\cite{Perez2013-AutomaticModelInterestingness} proposed a model to evaluate interestingness based on novelty and correct story recount, with emphasis on the story's opening, closure, and dramatic tensions. Szilas et al.~\cite{Szilas2016-QualQuantInterestingness} discuss interestingness as a paradox dramatic situation with obstacles and conflicts, albeit applicable to stories as successive events. Yet, to approach subjective measurements such as interestingness, most research turns towards having human evaluation~\cite{Kreminski2019-StorySifter,Lankoski2013-StoryConsistencyInteresting} or using such to form human models to be used as surrogate models~\cite{Sharna2007-PreferenceModelingStories,Li2010-plannerPlotAdapt}.



\begin{table}[]
\caption{Tropes included and used in TropeTwist, extracted from~\cite{tvtropes}.}
\resizebox{\linewidth}{!}{%
\begin{tabular}{l|l|l}
Name                       & \textit{Symbol} & \textit{Definition}                                                                  \\ \hline
Hero                       & HERO            & A protagonist character.                                                             \\
Five-man band              & 5MA             & \begin{tabular}[t]{@{}l@{}}Group composed by up-to-five \\ archetypical characters.\end{tabular}                            \\
The chosen one             & NEO             & Specific hero chosen as the one.                                                     \\
Superhero                  & SH              & Specific hero with unique abilities.                                                 \\
Conflict                   & CONF             & \begin{tabular}[t]{@{}l@{}}Non-specific problem to overcome \\ between characters.\end{tabular}                                   \\
Enemy                      & ENEMY           & A nemesis to the hero.                                                               \\
Empire                     & EMP             & \begin{tabular}[t]{@{}l@{}}Collective enemy with the \\ambition of conquering the world.\end{tabular}                               \\
Big bad                    & BAD             & \begin{tabular}[t]{@{}l@{}}Specific enemy, which is the \\ ultimate cause for all the bad.\end{tabular}                               \\
Dragon                     & DRAKE             & \begin{tabular}[t]{@{}l@{}}Specific enemy, which is the right \\ hand of BAD.  \end{tabular}                                          \\
Plot device                & PLD             & \begin{tabular}[t]{@{}l@{}}A feature or element that drives \\ the plot forward. \end{tabular}                                        \\
Chekhov's gun              & CHK             & PLD relevant to the story \\
MacGuffin                  & MCG             & \begin{tabular}[t]{@{}l@{}}PLD with irrelevant nature to \\ drive the story.\end{tabular}          \\
May help in quest & MHQ             & PLD important to resolve a conflict.    
\end{tabular}%
}
\label{tab:tropes}
\end{table}

\section{Building narrative structures with tropes}

In storytelling, a trope~\cite{tropesSimpsons} is a convention or figure of speech that the storyteller assumes to be recognizable by the audience. TvTropes is an online wiki that compiles and describes several thousand tropes in many sorts of media~\cite{tvtropes}. As exemplified by \cite{periodicTable}, tropes could be interconnected in graph-like structures, called story molecules, to succinctly depict the structure behind a narrative. 

\subsection{TropeTwist}

TropeTwist elaborates on the concept of story molecule to represent narratives using graph-like structures of interconnected tropes, called narrative graphs (NG). NGs encode narrative structures in an abstract level that show and define the game's narrative structure and certain abstract properties such as key items, roles, relations, or main events. Table \ref{tab:tropes} shows all the included tropes to be used as nodes. Nodes are depicted (fig~\ref{fig:teaserfig}) with shapes specific to their trope base type: heroes (rectangle), conflicts (diamond), enemies (hexagon), and plot devices (circle). HERO is the base pattern of 5MA, NEO, and SH. ENEMY is the base pattern of EMP, BAD, and DRAKE. PLD is the base pattern of CHK, MCG, and MHQ.

Nodes in a narrative graph are necessarily interconnected by either unidirectional or bidirectional edges (with one or both arrowheads) or by entailment edges (with a single diamond head). Given nodes A and B, A $\diamondsuit$--- B, reads as ``A entails B,'' whereas A $\rightarrow$ B denotes a relationship from A to B, and B $\rightarrow$ A the opposite. A $\leftrightarrow$ B denotes a reflexive relationship between A and B. As an example, HERO $\rightarrow$ CONFLICT $\rightarrow$ EMP denotes a hero who is in conflict against an empire-type enemy, whereas HERO $\leftrightarrow$ CONFLICT denotes a hero who is in conflict with themselves. EMP $\diamondsuit$--- DRAKE $\diamondsuit$--- NEO, denotes an empire that entails a dragon enemy that, once beaten, will lead to the appearance of a chosen one hero, creating some causal links. The system is ambiguous by design. We take advantage of the ambiguity for 1) the generation of new structures (fewer constraints), 2) removing the focus on details by designers to let them focus on the overarching picture, and 3) for other systems to define and interpret these abstract properties. 

Furthermore, interconnecting tropes can give rise to other tropes and patterns, described in the following section. The nodes and their respective trope and pattern were chosen from a subset of tropes in generic categories such as heroes or plot devices. These categories were inspired and chosen based on tropes from TVTropes, the division by James Harris~\cite{periodicTable}, and previous research such as Propp's morphology~\cite{propp1975-morphology} or Greimas' actantial model~\cite{Greimas84-structuralSemantics}. 

\subsection{Trope Patterns}


Tropes and interconnected tropes (i.e., subgraphs) give rise to different types of patterns. These patterns can be \textbf{micro-patterns}, encapsulating a single trope node, \textbf{meso-patterns}, often composed by more than one micro-pattern with special meaning, and \textbf{auxiliary patterns}, denoting graph problems. We calculate the relative tropes and patterns' quality within an NG and use this to assess the general quality of the graph. These qualities are proxies for certain characteristics among the defined patterns that are used to evaluate the graphs, but they do not capture any story quality; especially, since we are only defining structures. When generating narrative graphs from a root (explained in section~\ref{sec:evolvingNarratives}), the quality of a narrative graph becomes relative to the root, henceforth, the ``root graph'' (RG). In the following descriptions, we will use \textbf{EG} referring to the ``evaluated graph'' we are calculating the pattern's quality (the generated individual), and \textbf{RG} to refer to the relative and root graph. When using subscript ``pat,'' we refer to the current pattern that is evaluated. 


For most patterns, we calculate three general qualities (indicated when used) that add to the quality of the pattern. $G_{q}(pattern)$ relates to the \textit{Generic} quality of patterns in EG, which calculates the general occurrence of a pattern within EG compared to its occurrence in RG, calculated in eq.~\ref{eq:generic_qual}. $R_{q}(pattern)$ relates to the \textit{Repetition} quality of patterns, which calculates if a trope is unique in EG ($R_{q}(pattern) = 1$) or its ratio among the same base pattern. Lastly, $I_{q}(pattern)$ relates to the \textit{Involvement} quality of patterns in EG, which calculates the amount of associations a pattern has with \textbf{structure patterns}. Involvement means that the pattern is either \textit{source} or \textit{target} in a structure and is calculated as the ratio of structure pattern involvement by the structure pattern count in EG. These three metrics incentivize graphs with similar amount and type of nodes than RG, minimal repetitions, and more involvement. 

\begin{equation}
\label{eq:generic_qual}
    G_{q}(pattern) = 1.0 - |RG_{pat} - EG_{pat}|/\max(RG_{pat}, EG_{pat})
\end{equation}




\subsubsection{Micro-Patterns}

Micro-patterns are the fundamental unit in the system, which aims at categorizing different sets of the individual patterns that are shown in table~\ref{tab:tropes}. Micro-patterns are single nodes and the basic building block that, when interconnected, allows the detection of meso-patterns.

\emph{Structure Pattern (SP)} is any type of trope that would give some structural definition to a narrative, whether this being a conflict, specific act, or a part in a dramatic arc (e.g., climax). Currently, the only type of structure trope is the \textsc{conflict} (CONF) trope, which represents the most basic structural interaction. The quality $SP_{q}$ is calculated as the equally weighted linear combination of:

\begin{equation}
    SP_{q} = G_{q}(SP) + I_{q}(SP)
\end{equation}

\emph{Character Pattern (CP):} are identified as nodes within the narrative that could be either the player, possible ally or enemy NPCs, or simple enemies. In TropeTwist, it is distinguished between heroes and villain patterns, and these are commonly used as \textbf{sources} or \textbf{targets} (or both) of other patterns, and on a few special occasions to denote a relation to another character. The quality $CP_{q}$ is calculated per group (heroes and villains), and it is the equally weighted linear combination of:

\begin{equation}
    CP_{q} = G_{q}(CP) + R_{q}(CP) + I_{q}(CP)
\end{equation}

\emph{Plot Device Pattern (PDP)} is described as the element within the narrative that moves it forward, as a goal, object, or dramatic element to be used or encountered by any of the characters. The quality $PDP_{q}$ is calculated as the equally weighted linear combination of:

\begin{equation}
    PDP_{q} = G_{q}(PDP) + R_{q}(PDP)
\end{equation}

\subsubsection{Meso-Patterns}
Meso-patterns are the features that emerge in the narrative from dynamically combining micro-patterns and, on some occasions, these with other meso-patterns. They are always composed of more than one pattern denoting some spatial, semantic, or usability relationship within the narrative graph. We identified a subset of Tropes (extracted from TVTropes~\cite{tvtropes}) that requires or works as the combination between more fundamental units. For instance, the \textit{reveal pattern} relates to the ``Good all along'' or ``evil all along.''

\emph{Conflict Pattern (ConfP)} is a type of structure pattern composed by a conflict node (Con), a source $s$ node, and a target $t$ node, which are both CPs and usually a hero and a villain or the same character as $s$ and $t$. For instance, the subgraph HERO $\rightarrow$ CONFLICT $\rightarrow$ EMP, indicates that a hero CP has a conflict with an enemy CP. A conflict node can be used indefinitely to define several ConfP. A ConfP is also either~\textsc{explicit} or~\textsc{implicit}. \textsc{Explicit} conflicts are explicitly encoded in the graph and directed from $s$ to $t$ passing through the conflict trope. On the other hand, \textsc{Implicit} conflicts relates to the conflicts from $t$ (or derivatives) to $s$ (or derivatives) that are not encoded in the graph. For instance, the previous example is an \textsc{explicit} conflict from HERO to EMP, and at the same, the EMP has an \textsc{implicit} conflict with the HERO. The quality $ConfP_{q}$ is calculated as the equally weighted linear combination of:

\begin{equation}
    ConfP_{q} = G_{q}(ConfP) + R_{q}(ConfP)
\end{equation}


\emph{Derivative Pattern (DerP)} defines a relationship between tropes connected by ``entails'' connections ($\diamondsuit$---). Therefore, a DerP contains a list of patterns connected by entails, named derivatives. DerP starts from a root micro-pattern and continue until no more ``entail'' connections are encountered, effectively establishing a hierarchy from the root derivative to the rest. By design, the patterns within a DerP have a local and temporal order and a causal relationship. For instance, in the subgraph EMP $\diamondsuit$--- DRAKE $\diamondsuit$--- NEO, engaging with the \emph{EMP}, entails both the conflict with \emph{DRAKE} and the appearance of \emph{NEO}. This means that only by overcoming the \emph{DRAKE}, \emph{NEO} will appear - as a new hero or the evolution of another. The quality $DerP_{q}$ is calculated (eq.~\ref{eq:derp}) based on its $G_{q}(DerP)$, the ratio of derivatives within the DerP among the total amount of derivatives across all DerPs in EG ($ratio\theta_{q}$), and the derivatives' diversity. 



\begin{equation}
\label{eq:derp}
    DerP_{q} = G_{q}(DerP) + 
    ratio\theta_{q} + 
    \frac{\sum_{i=0}^{len(DerP_{der})}DerP_{der_ibasepat}}{len(DerP_{der})}
\end{equation}

\emph{Reveal Pattern (RevP)} connects two independent CPs as one, meaning that character A was, in fact, always character B, and vice-versa. This pattern identifies confusion and surprise within an EG, as, for instance, a villain could have been, in fact, ``Good All Along''\footnote{https://tvtropes.org/pmwiki/pmwiki.php/Main/GoodAllAlong}. In practice, a RevP is identified as a villain or hero connected with a unidirectional connection ($\rightarrow$) to another hero or villain. As a consequence, all existing conflicts between them would become \emph{fake}. $RevP_{q}$ is calculated based on its $G_{q}(RevP)$, the number of reveals in EG in relation to characters, and the number of fake conflicts given the specific reveal.

\begin{multline}
    RevP_{q} = G_{q}(RevP) + \frac{len(EG_{RevP})}{len(EG_{CP})} +  \\ \Bigg( 1.0 - \frac{\sum_{i=0}^{len(EG_{conf})}    \begin{cases}
        1,& \text{if } RevP \in x_{i}\\
        0,              & \text{otherwise}
    \end{cases}}{len(EG_{conf})} \Biggl)
\end{multline}

\emph{Active Plot Device Pattern (APD)} operationalize and integrate PDPs within a narrative since PDPs only describe an abstract goal or target. In practice, an \textit{APD} is identified as PDPs that have at least one incoming connection, and optionally, one single outgoing connection. These limitations are added to limit the effect of a PDP within a narrative. $APD_{q}$ is measured based on its $G_{q}(APD)$, and the APD's usability, calculated based on the sum of incoming and outgoing connections divided by half of the nodes in EG depicted as $bal\gamma_{q}$, penalizing APDs for not using all their connections. 

\begin{equation}
    APD_{q} = G_{q}(APD) + bal\gamma_{q}
\end{equation}

\emph{Plot Points (PP)} are key events within the EG, identified as discrete moments given some pattern. The derivatives within a \textit{DerP}, RevP's source, and PDPs that are \textit{APD} are considered as plot points. $PP_{q}$ is measured based on the number of PPs within RG ($ G_{q}(PP)$), and the number of PPs within EG in relation to the number nodes within it ($Balance_{q}(PP)$).

\begin{equation}
    PP_{q} = G_{q}(PP) + Balance_{q}(PP)
\end{equation}

\emph{Plot Twist (PT)} takes advantage of plot points to identify those that could have a bigger impact on the narrative. In practice, \emph{PTs} consider the source of \textit{RevP}, derivatives from \textit{DerP} that are a different micro-pattern than the root of the DerP (except PDPs), and \textit{APDs} that are connected to other \textit{APDs}. For instance, in the subgraph: EMP $\diamondsuit$--- DRAKE $\diamondsuit$--- NEO, given that NEO is a different micro-pattern than root EMP (Hero and Villain, respectively), NEO will be identified as a \textit{Plot Twist} as it alters the ``natural'' order in the DerP. $PT_{q}$ is based on the number of PTs within RG ($ G_{q}(PT)$), the PT's involvement in EG, and the balance of PTs based on the PPs in EG. Involvement varies depending on the associated pattern to PT. When a PT is associated with a \textit{RevP}, involvement is calculated as how much the structure changes based on that (i.e., how many fake conflicts are created). When it is related to \textit{DerP}, involvement is calculated as how different the pattern is and its order within the derivatives. Finally, when it is related to \textit{APD}, involvement is based on how usable the \textit{APD} is within the narrative based on incoming and outgoing connections.

\begin{equation}
    PT_{q} = G_{q}(PT) + I_{q}(PT_{assoc_pat}) + \frac{len(EG_{pt})}{len(EG_{pp})}
\end{equation}

\subsubsection{Auxiliary Patterns}

Auxiliary patterns denote problems in the graph and sub-optimal or impractical nodes and connections within a graph. They are classified into \textit{Nothing}, which are nodes that are not identified as part of a meso-pattern; and \textit{Broken Link}, which are outgoing connections from a node that are not used or do not lead to any pattern.

\subsection{Proof-of-Concept}
\label{sec:PoC}

TropeTwist can be used to represent different narrative structures and parts of games. To test and show TropeTwist's expressiveness, we chose to form three different narrative graphs representing different games shown in figure~\ref{fig:teaserfig}, top row: \emph{Zelda: Ocarina of Time} (Zelda:OoT)~\cite{tloz:oot}, \emph{Zelda: A Link to the Past} (Zelda:LttP)~\cite{tloz:lttp} - eastern palace, and \emph{Super Mario Bros} (SMB)~\cite{mario}. They represent different games from different genres (fig.~\ref{fig:teaserfig}.a and \ref{fig:teaserfig}.b are adventure-dungeon games, and \ref{fig:teaserfig}.c is a platformer), and represent different game's phases; in the case of fig.~\ref{fig:teaserfig}.a and \ref{fig:teaserfig}.c, both represent the main structure of the game, while \ref{fig:teaserfig}.b, represents a specific area and sequence of the game.

Figure~\ref{fig:teaserfig}.a represents a simplified overarching narrative structure from Zelda: OoT. The ocarina of time, given by Zelda to Link, is defined as a McGuffin (MCG) that, when collected by ``young link,'' allows him to go forward in time to ``adult link,'' the chosen one (NEO). This, in turn, enables explicit conflicts between hero and enemy characters, which represents the main loop of the game. The structure shows two factions, a set of heroes and the BAD. 


Figure~\ref{fig:teaserfig}.b represents the structure and plot points from the eastern palace in Zelda: LttP. All palaces in \textit{A Link to the Past} follow a very similar structure and sequence. The HERO's goal is to get the ``Pendant of Courage'' (MCG). However, the MCG derives from ENEMY and BAD, so the HERO must overcome them to achieve his goal. The structure shows a causal and linear narrative that could be used to identify elements that need to appear before others, similar to the work by Dormans and Bakkes~\cite{dormans2011generating}. 

Figure~\ref{fig:teaserfig}.c represents the overarching narrative structure of SMB. In SMB, the objective of Mario (HERO) is to rescue Peach (HERO) from Bowser (BAD). To do this, the player goes through a series of platform worlds that always end in a ``Fake Bowser'' (DRAKE). The player must continue until encountering the ``Real Bowser'' (BAD), which then would enable the player to get to their objective (MCG). 
\begin{figure*}[t]
    \centering
    \includegraphics[width=\textwidth]{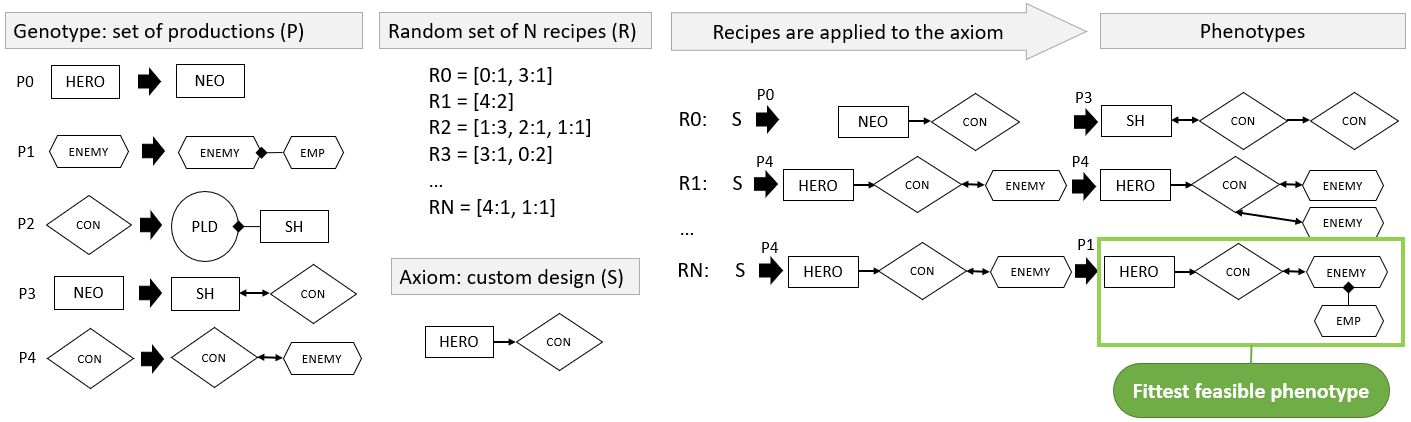}
    \caption{sample complete process from an individual's genotype to the phenotype.}
    \label{fig:gen2phen}
\end{figure*}

\section{Evolving Narratives with Graph Grammars} \label{sec:evolvingNarratives}

We use the Constrained MAP-Elites~\cite{Khalifa2018}, and adapt it to work with graph grammars, evolve production rules, and adapt the evolution towards a target similar to~\cite{Alvarez2020-ICMAPE}. Constrained MAP-Elites adds feasible-infeasible two populations to each cell, effectively evolving sub-populations per cell. An individual's phenotype is a narrative graph, and its encoding genotype is the production rules of a graph grammar. A graph grammar is a context-free grammar whose productions add, remove, and modify nodes and edges of a graph. Our implementation uses the tropes listed in Table \ref{tab:tropes} as nodes, and the three available connection types as edges ($\rightarrow$, $\leftrightarrow$, $\diamondsuit$--). Graph grammars do not apply rules sequentially; instead, every individual does a random sampling of the rules in their genotype to produce \emph{recipes} to generate graphs. \emph{Recipes} describe the rules' order and repetition, and their size is limited by the amount of production rules as minimum and the minimum plus five as maximum. \emph{Recipes} do not have repetitions within them, i.e., if rule 1 is added at step 2, subsequent addition would simply add to the number of times that rule will be applied at step 2. Their size is limited by the number of production rules as minimum and up to five more samples as maximum. Figure~\ref{fig:gen2phen} shows a sample complete process from an individual's genotype (i.e., rules) to the phenotype (i.e., narrative graph).


Individuals move between the feasible and infeasible population depending on the feasibility constraint. NGs are deemed infeasible if the nodes are not fully connected or if there exists a conflict pattern with more than one self-conflict. Infeasible individuals are evaluated based on how close they are to be fully connected and not having any inadequate self-conflict. The fitness function assesses NGs that are deemed feasible based on their coherence (equation~\ref{eq:coherence_fitness}), which we use to assess how correct, coherent, and in general, syntactically correct the narrative graphs are. Coherence aims at maximizing an equally weighted sum between cohesion and consistency. Cohesion refers to the link between elements that hold together to form some group. In our implementation, it focuses on minimizing the number of auxiliary patterns by calculating the proportion of \emph{Nothing} and \emph{Broken Link} among all patterns in NG. A consistent NG should be regular and free of contradictions. Thus, we calculate \emph{consistency} (eq.~\ref{eq:consistency_fitness}) as the collective quality of micro-patterns since they are the building blocks, and conflicts' goodness based on the number of fake conflicts. Thus, we aim at maximizing the quality of micro-patterns and minimizing contradictions created by meso-patterns.



\begin{equation}
\label{eq:consistency_fitness}
f_{consistency} = \frac{\sum_{i=0}^{len(ng_{micro})} i_{qual}}{len(ng_{micropat})} -  \\ 
\frac{len(ng_{fakeConfP})}{len(ng_{confP})} 
\end{equation}

\begin{equation}
\label{eq:coherence_fitness}
f_{coherence} = f_{consistency} + (1.0 - f_{cohesion})
\end{equation}

Furthermore, MAP-Elites uses behavioral dimensions in a grid shape to retain and foster diversity throughout generations. We use the following two dimensions to evaluate the diversity: 

\textbf{Step.} Step (eq.~\ref{eq:StepDim}) calculates the Levenshtein distance~\cite{Levenshtein96-editDistance} between two narrative graphs, taking into consideration the number and type of nodes and connections. Step is normalized using step threshold $\theta = 11$ determined through a process of experimentation, which does not consider steps farther than $\theta$, avoiding the generation of too dissimilar graphs.

\begin{equation}
\label{eq:StepDim}
D_{step} =  \min (lev_{EG,RG} (|EG|, |RG|), \theta)
\end{equation}

\textbf{Interestingness (int).} We aim at measuring the semantic quality of a narrative graph. A narrative graph can be syntactically correct and coherent yet lack a good semantic quality and do not evoke interest for designers or players. Therefore, we leverage \textbf{plot point}, \textbf{plot twist}, and \textbf{active plot device} patterns to measure the \emph{interestingness} of the NGs. The nature of \emph{interestingness} creates pressure on the fitness function since the incidence of the three meso-patterns could (if overused) ``degenerate'' the narrative; thus, decreasing its coherence. $D_{int}$ is calculated as the weighted sum ($w_{0}=0.4, w_{1}=0.2, w_{2}=0.4$) of the normalized cumulative quality of \textbf{APDs}, \textbf{PPs}, and \textbf{PTs} within an NG (eq.~\ref{eq:interesting_fitness}).

\begin{equation}
\label{eq:interesting_fitness}
D_{int} = w_{0} \times \frac{APD_{q}}{\#APD} + w_{1} \times \frac{\#PP_{q}}{\#PP} +  w_{2} \times \frac{PT_{q}}{\#PT}
\end{equation}







\subsection{Experiments}




\begin{table}[t]
\caption{Comparative results between root graphs and generated elites (shown in fig. \ref{fig:teaserfig})}
\resizebox{\columnwidth}{!}{%
\begin{tabular}{|l|c|c|c|c|}
\hline
Graph                        & \multicolumn{1}{l|}{Cohesion} & \multicolumn{1}{l|}{Consistency} & \multicolumn{1}{l|}{Coherence (fitness)} & \multicolumn{1}{l|}{Interestingness} \\ \hline
RG (fig \ref{fig:teaserfig}.a1)    & 1.0                           & 0.66                             & 0.825                                     & 0.61                                 \\
Elite (fig \ref{fig:teaserfig}.a2) & 1.0                           & 0.76                             & 0.875                                     & 0.73                                 \\ \hline
RG (fig \ref{fig:teaserfig}.b1)    & 1.0                           & 0.75                             & 0.87                                     & 0.38                                 \\
Elite (fig \ref{fig:teaserfig}.b2) & 1.0                           & 0.91                             & 0.95                                    & 0.55                                 \\ \hline
RG (fig \ref{fig:teaserfig}.c1)    & 1.0                           & 0.77                             & 0.88                                     & 0.4                                  \\
Elite (fig \ref{fig:teaserfig}.c2) & 1.0                           & 0.85                             & 0.92                                     & 0.52                                 \\ \hline
\end{tabular}
}
\label{tab:best-generated}
\end{table}

We conducted a series of experiments to evaluate and analyze how the system could evolve NGs into quality-diverse and valid narrative structures. We evolved the three manually constructed narrative graphs shown in figure~\ref{fig:teaserfig}, top row. They were used as root graphs and axioms in the EA, and we used \textit{interestingness} and \textit{step} as behavioral dimensions. We did $5$ MAP-Elites runs per narrative graph, ran each for $500$ generations, and set the initial population to $1000$ randomly created individuals. The initial population is generated by randomly creating between two and five production rules. Each feasible and infeasible population per cell has $25$ individuals. Each individual is limited to test $10$ recipes regardless of the chromosome size. Offspring were produced either by selecting either the left-side or right-side of a random production rule and exchanging them or with a $50$\% mutation chance. If an offspring was generated by mutation, there was a $10$\% chance to add or remove a production rule and a $90$\% to modify in various ways existing production rules.


We calculated the \textit{coverage}: how much of the constrained search space is explored (i.e., constrained by the behavioral dimensions); the avg. fitness and the avg. interestingness. All experiments had little variation regarding these metrics, and got in avg. 23.5\% coverage (24.9\%, 21.4\%, and 24.2\%, respectively), 0.79 fitness (0.76, 0.8, 0.8, respectively), and 0.37 interestingness (0.39, 0.37, 0.36, respectively). These results exemplify both the hard task of generating narrative graphs and exploring the possibility space, and the seemingly competing qualities of coherence (i.e., fitness) and interestingness. 



Furthermore, in figure~\ref{fig:teaserfig}, bottom row, it is shown three different example elite narrative graphs, generated from their respective root graphs on the top row and with each individual evaluation shown in table~\ref{tab:best-generated}. The root graphs have a cohesion of 1.0 since none of them have unused nodes or connections and have similar mid-high consistency values because of using generic nodes (e.g., HERO or ENEMY), repeating them, and low involvement in structures by characters. In the case of fig \ref{fig:teaserfig}.a1, the \textbf{RevP} from HERO to SH creates some fake conflicts, which affect the consistency but also boost the interestingness value of the narrative graph. Both fig \ref{fig:teaserfig}.b1 and \ref{fig:teaserfig}.c1, are evaluated similarly with low interestingness; c1 involves a simplistic and linear structure, and b1, while in principle more complex, is also a relatively linear structure with no \textbf{PTs}.



Furthermore, all the exemplar elites have better \textit{consistency}, \textit{coherence}, and \textit{interestingness} than the respective root graph. In figure~\ref{fig:teaserfig}.a2, the graph has been reduced towards a bottleneck, \textbf{RevP} (HERO $\rightarrow$ SH) is removed, and MCG is added as the objective for SH, which could point towards competition or cooperation to enable NEO. Such a change gives more \textit{consistency} to the graph while seemingly reducing its \textit{interestingness}, but this relation and the $\diamondsuit$-- connection between MCG and NEO increase its \textit{interestingness}. In figure~\ref{fig:teaserfig}.b2, the narrative has more interaction between \textbf{Plot Devices}, and the BAD has a more active role. Particularly, the fact that now HERO $\rightarrow$ MCG $\diamondsuit$-- BAD and MHQ $\diamondsuit$-- CHK $\diamondsuit$-- BAD could enable and force the HERO towards two main objectives before overcoming the boss, which is reflected in the higher \textit{Interestingness}. Finally, in figure~\ref{fig:teaserfig}.c2, the narrative did not change much (only four steps away), yet the graph is seemingly better, and the narrative could be very different. The graph has broken the loop which connected DRAKE $\diamondsuit$-- BAD, and could could point towards a side objective. Further, the connection between BAD and MCG has been reversed; thus, the HERO does not need to face the BAD to get the MCG, rather reaching the MCG will have as a consequence the emergence of the BAD. Finally, BAD is no longer connected to EMP and DRAKE; thus, BAD could be its own enemy faction, in this case, complexifying the narrative and creating more challenge.
\section{Discussion and Limitations}




The trope-graph representation in TropeTwist allows for a quick definition of narrative structures. They are, by design, ambiguous, do not encode temporal information besides causal chains, and are, to some extent, generic, which makes structures relatively simple to develop but more complex to interpret. These design decisions make the system encode less rich information than others, such as Scheherazade~\cite{elson-2012-dramabank}, but allow the structure to be interpreted in multiple ways. For instance, the generated graphs could equally describe different stories, and the interpretation given in this paper is just one of many. Thus, the system effectively shifts the complexity from the structure to the ``interpreter.'' While the generated structures could already serve as inspiration for users, an interpreter could provide alternative interpretations that could be guided by or learned from users, which is part of our future work.

Furthermore, the metrics proposed and developed here were used to tune and evaluate the graph outputs without humans in the loop. However, they do not stand in or replace human judgment. The metrics are estimated heuristics mainly based on the graph functionality and relation among patterns. Most of them are related to a ``root graph,'' which is a preliminary step for making TropeTwist interactive and have humans-in-the-loop. We aim to develop a mixed-initiative version of TropeTwist, where metrics depend on the designer's creation. This would, in turn, allow the designer to steer the MAP-Elites search, generating content adapted to them~\cite{alvarez_assessing_2021}, and for MAP-Elites to assist designers with ideation proposing varied structures.


 
\section{Conclusions and Future Work}


In this paper, we have presented \emph{TropeTwist}, a system that interconnects tropes and trope patterns to describe narrative structures. We demonstrated through three proof-of-concept structures the system's expressiveness to describe games with diverse genres and mechanics, and different game phases. Further, we illustrated how we could generate novel structures from the three proof-of-concept structures using MAP-Elites, improving them on our metrics. 

Tropes could be seem as something to avoid when exploring creativity, mainly due to the possibility of showing unoriginal views by definition. However, a set of combined tropes, patterns, and structures could give rise to novel combinations that express the wanted structure. Similarly, identifying, visualizing, and defining the tropes and patterns and doing ``twists'' with them; thus, transforming something typical into atypical is the goal with TropeTwist.

The narrative structures show essential aspects of how the story will develop and lead, and important components such as events, conflicts, or roles. However, to further operationalize these structures, it is necessary other systems that make use of them, such as quest~\cite{Alvarez2021-questgram,ammanabrolu2019-towardQuestGeneration} or plot~\cite{Ammanabrolu2020-PlotEventsSentences} generators. Another interesting future work would be to explore the multi-faceted nature of games~\cite{Liapis2019-OrchestratingGames} and combine this type of system with generators that focus on other facets such as level design~\cite{sarkar2021-dungeonPlatformer,alvarez2019empowering} or game mechanics~\cite{green2021-gamemechanicsAlignment,charity2020mech}.

Generating novel narrative structures resulted in interesting variations, but the system could not exploit all the advantages of MAP-Elites. Our results point towards difficulties exploring the space, possibly because \emph{coherence} and \emph{interestingness} are to some extent competing objectives. Therefore, we aim at extending TropeTwist towards a mixed-initiative co-creative system~\cite{yannakakis2014micc}, and with that, evaluate with human participants. Given that our metrics are dependant on the designed graph; then, we could constantly adapt the content generation and have adaptive models, for instance, of interestingness, based on the user's creation similar to~\cite{alvarez2019empowering,Panagiotis2021-susketch}.

\bibliographystyle{ACM-Reference-Format}
\bibliography{references,games}


\begin{thebibliography}{62}


\ifx \showCODEN    \undefined \def \showCODEN     #1{\unskip}     \fi
\ifx \showDOI      \undefined \def \showDOI       #1{#1}\fi
\ifx \showISBNx    \undefined \def \showISBNx     #1{\unskip}     \fi
\ifx \showISBNxiii \undefined \def \showISBNxiii  #1{\unskip}     \fi
\ifx \showISSN     \undefined \def \showISSN      #1{\unskip}     \fi
\ifx \showLCCN     \undefined \def \showLCCN      #1{\unskip}     \fi
\ifx \shownote     \undefined \def \shownote      #1{#1}          \fi
\ifx \showarticletitle \undefined \def \showarticletitle #1{#1}   \fi
\ifx \showURL      \undefined \def \showURL       {\relax}        \fi
\providecommand\bibfield[2]{#2}
\providecommand\bibinfo[2]{#2}
\providecommand\natexlab[1]{#1}
\providecommand\showeprint[2][]{arXiv:#2}

\bibitem[Aarseth(2005)]%
        {aarseth2005hunt}
\bibfield{author}{\bibinfo{person}{Espen Aarseth}.}
  \bibinfo{year}{2005}\natexlab{}.
\newblock \showarticletitle{From hunt the wumpus to everquest: introduction to
  quest theory}. In \bibinfo{booktitle}{\emph{International Conference on
  Entertainment Computing}}. Springer, \bibinfo{pages}{496--506}.
\newblock


\bibitem[Aarseth(2012)]%
        {Aarseth2012-Narrativetheory}
\bibfield{author}{\bibinfo{person}{Espen Aarseth}.}
  \bibinfo{year}{2012}\natexlab{}.
\newblock \showarticletitle{A Narrative Theory of Games}. In
  \bibinfo{booktitle}{\emph{Proceedings of the International Conference on the
  Foundations of Digital Games}} (Raleigh, North Carolina)
  \emph{(\bibinfo{series}{FDG '12})}. \bibinfo{publisher}{ACM},
  \bibinfo{address}{New York, NY, USA}, \bibinfo{pages}{129–133}.
\newblock
\showISBNx{9781450313339}
\urldef\tempurl%
\url{https://doi.org/10.1145/2282338.2282365}
\showDOI{\tempurl}


\bibitem[Aarseth and Grabarczyk(2018)]%
        {aarseth2018-ontologicalMeta}
\bibfield{author}{\bibinfo{person}{Espen Aarseth} {and} \bibinfo{person}{Paweł
  Grabarczyk}.} \bibinfo{year}{2018}\natexlab{}.
\newblock \showarticletitle{An Ontological Meta-Model for Game Research}. In
  \bibinfo{booktitle}{\emph{DiGRA \&\#3918 - Proceedings of the 2018 DiGRA
  International Conference: The Game is the Message}}.
  \bibinfo{publisher}{DiGRA}.
\newblock
\urldef\tempurl%
\url{http://www.digra.org/wp-content/uploads/digital-library/DIGRA_2018_paper_247_rev.pdf}
\showURL{%
\tempurl}


\bibitem[Alvarez et~al\mbox{.}(2019)]%
        {alvarez2019empowering}
\bibfield{author}{\bibinfo{person}{Alberto Alvarez}, \bibinfo{person}{Steve
  Dahlskog}, \bibinfo{person}{Jose Font}, {and} \bibinfo{person}{Julian
  Togelius}.} \bibinfo{year}{2019}\natexlab{}.
\newblock \showarticletitle{Empowering Quality Diversity in Dungeon Design with
  Interactive Constrained MAP-Elites}. In \bibinfo{booktitle}{\emph{2019 IEEE
  Conference on Games (CoG)}}.
\newblock


\bibitem[Alvarez et~al\mbox{.}(2020)]%
        {Alvarez2020-ICMAPE}
\bibfield{author}{\bibinfo{person}{Alberto Alvarez}, \bibinfo{person}{Steve
  Dahlskog}, \bibinfo{person}{Jose Font}, {and} \bibinfo{person}{Julian
  Togelius}.} \bibinfo{year}{2020}\natexlab{}.
\newblock \showarticletitle{Interactive Constrained MAP-Elites: Analysis and
  Evaluation of the Expressiveness of the Feature Dimensions}.
\newblock \bibinfo{journal}{\emph{IEEE Transactions on Games}}
  (\bibinfo{year}{2020}).
\newblock


\bibitem[Alvarez et~al\mbox{.}(2021a)]%
        {alvarez_assessing_2021}
\bibfield{author}{\bibinfo{person}{Alberto Alvarez}, \bibinfo{person}{Jose
  Font}, \bibinfo{person}{Steve Dahlskog}, {and} \bibinfo{person}{Julian
  Togelius}.} \bibinfo{year}{2021}\natexlab{a}.
\newblock \showarticletitle{Assessing the {Effects} of {Interacting} with
  {MAP}-{Elites}}. In \bibinfo{booktitle}{\emph{Proceedings of the {AAAI}
  {Conference} on {Artificial} {Intelligence} and {Interactive} {Digital}
  {Entertainment}}}, Vol.~\bibinfo{volume}{17}. \bibinfo{publisher}{AAAI},
  \bibinfo{pages}{124--131}.
\newblock
\urldef\tempurl%
\url{https://ojs.aaai.org/index.php/AIIDE/article/view/18899}
\showURL{%
\tempurl}


\bibitem[Alvarez et~al\mbox{.}(2021b)]%
        {Alvarez2021-questgram}
\bibfield{author}{\bibinfo{person}{Alberto Alvarez}, \bibinfo{person}{Eric
  Grevillius}, \bibinfo{person}{Elin Olsson}, {and} \bibinfo{person}{Jose
  Font}.} \bibinfo{year}{2021}\natexlab{b}.
\newblock \showarticletitle{Questgram [Qg]: Toward a Mixed-Initiative Quest
  Generation Tool}. In \bibinfo{booktitle}{\emph{Proceedings of the 16th
  International Conference on the Foundations of Digital Games}}
  \emph{(\bibinfo{series}{FDG'21})}. \bibinfo{publisher}{Association for
  Computing Machinery}, \bibinfo{address}{New York, NY, USA}.
\newblock


\bibitem[Ammanabrolu et~al\mbox{.}(2019)]%
        {ammanabrolu2019-towardQuestGeneration}
\bibfield{author}{\bibinfo{person}{Prithviraj Ammanabrolu},
  \bibinfo{person}{William Broniec}, \bibinfo{person}{Alex Mueller},
  \bibinfo{person}{Jeremy Paul}, {and} \bibinfo{person}{Mark Riedl}.}
  \bibinfo{year}{2019}\natexlab{}.
\newblock \showarticletitle{Toward Automated Quest Generation in Text-Adventure
  Games}. In \bibinfo{booktitle}{\emph{Proceedings of the 4th Workshop on
  Computational Creativity in Language Generation}}.
  \bibinfo{publisher}{Association for Computational Linguistics},
  \bibinfo{address}{Tokyo, Japan}, \bibinfo{pages}{1--12}.
\newblock
\urldef\tempurl%
\url{https://www.aclweb.org/anthology/2019.ccnlg-1.1}
\showURL{%
\tempurl}


\bibitem[Ammanabrolu et~al\mbox{.}(2020)]%
        {Ammanabrolu2020-PlotEventsSentences}
\bibfield{author}{\bibinfo{person}{Prithviraj Ammanabrolu},
  \bibinfo{person}{Ethan Tien}, \bibinfo{person}{Wesley Cheung},
  \bibinfo{person}{Zhaochen Luo}, \bibinfo{person}{William Ma},
  \bibinfo{person}{Lara~J. Martin}, {and} \bibinfo{person}{Mark~O. Riedl}.}
  \bibinfo{year}{2020}\natexlab{}.
\newblock \showarticletitle{Story Realization: Expanding Plot Events into
  Sentences}. In \bibinfo{booktitle}{\emph{The Thirty-Fourth {AAAI} Conference
  on Artificial Intelligence, {AAAI} 2020, The Thirty-Second Innovative
  Applications of Artificial Intelligence Conference, {IAAI} 2020, The Tenth
  {AAAI} Symposium on Educational Advances in Artificial Intelligence, {EAAI}
  2020, New York, NY, USA, February 7-12, 2020}}. \bibinfo{publisher}{{AAAI}
  Press}, \bibinfo{pages}{7375--7382}.
\newblock
\urldef\tempurl%
\url{https://aaai.org/ojs/index.php/AAAI/article/view/6232}
\showURL{%
\tempurl}


\bibitem[Baikadi and Cardona-Rivera(2012)]%
        {Baikadi2012-Narreme}
\bibfield{author}{\bibinfo{person}{Alok Baikadi} {and}
  \bibinfo{person}{Rogelio~E. Cardona-Rivera}.}
  \bibinfo{year}{2012}\natexlab{}.
\newblock \showarticletitle{Towards finding the fundamental unit of narrative:
  A Proposal for the Narreme}. In \bibinfo{booktitle}{\emph{Proceedings of the
  Third Workshop on Computational Models of Narrative}}.
  \bibinfo{publisher}{European Language Resource Association},
  \bibinfo{pages}{42--44}.
\newblock
\urldef\tempurl%
\url{https://doi.org/10.1.1.487.7222}
\showDOI{\tempurl}


\bibitem[Barthes(1975)]%
        {Barthes75-introStructNarr}
\bibfield{author}{\bibinfo{person}{Roland Barthes}.}
  \bibinfo{year}{1975}\natexlab{}.
\newblock \showarticletitle{An Introduction to the Structural Analysis of
  Narrative}.
\newblock \bibinfo{journal}{\emph{New Literary History}} \bibinfo{volume}{6},
  \bibinfo{number}{2} (\bibinfo{year}{1975}), \bibinfo{pages}{237--272}.
\newblock
\showISSN{00286087, 1080661X}
\urldef\tempurl%
\url{http://www.jstor.org/stable/468419}
\showURL{%
\tempurl}
\newblock
\shownote{Translated by Lionel Duisit}.


\bibitem[Blow(2004)]%
        {Blow2004-gamesHard}
\bibfield{author}{\bibinfo{person}{Jonathan Blow}.}
  \bibinfo{year}{2004}\natexlab{}.
\newblock \showarticletitle{Game Development: Harder Than You Think}.
\newblock \bibinfo{journal}{\emph{Queue}}  \bibinfo{volume}{1}
  (\bibinfo{date}{02} \bibinfo{year}{2004}), \bibinfo{pages}{28--37}.
\newblock
\urldef\tempurl%
\url{https://doi.org/10.1145/971564.971590}
\showDOI{\tempurl}


\bibitem[Castricato et~al\mbox{.}(2021)]%
        {Castricato2021-FabulaStoryCoherenceMeasure}
\bibfield{author}{\bibinfo{person}{Louis Castricato}, \bibinfo{person}{Spencer
  Frazier}, \bibinfo{person}{Jonathan~C. Balloch}, {and} \bibinfo{person}{Mark
  Riedl}.} \bibinfo{year}{2021}\natexlab{}.
\newblock \showarticletitle{Fabula Entropy Indexing: Objective Measures of
  Story Coherence}.
\newblock \bibinfo{journal}{\emph{CoRR}}  \bibinfo{volume}{abs/2104.07472}
  (\bibinfo{year}{2021}).
\newblock
\showeprint[arxiv]{2104.07472}
\urldef\tempurl%
\url{https://arxiv.org/abs/2104.07472}
\showURL{%
\tempurl}


\bibitem[Charity et~al\mbox{.}(2020a)]%
        {charity2020mech}
\bibfield{author}{\bibinfo{person}{Megan Charity},
  \bibinfo{person}{Michael~Cerny Green}, \bibinfo{person}{Ahmed Khalifa}, {and}
  \bibinfo{person}{Julian Togelius}.} \bibinfo{year}{2020}\natexlab{a}.
\newblock \showarticletitle{Mech-Elites: Illuminating the Mechanic Space of
  GVG-AI}. In \bibinfo{booktitle}{\emph{International Conference on the
  Foundations of Digital Games}} (Bugibba, Malta) \emph{(\bibinfo{series}{FDG
  '20})}. \bibinfo{publisher}{Association for Computing Machinery},
  \bibinfo{address}{New York, NY, USA}, Article \bibinfo{articleno}{8},
  \bibinfo{numpages}{10}~pages.
\newblock
\showISBNx{9781450388078}
\urldef\tempurl%
\url{https://doi.org/10.1145/3402942.3402954}
\showDOI{\tempurl}


\bibitem[Charity et~al\mbox{.}(2020b)]%
        {charity2020baba}
\bibfield{author}{\bibinfo{person}{Megan Charity}, \bibinfo{person}{Ahmed
  Khalifa}, {and} \bibinfo{person}{Julian Togelius}.}
  \bibinfo{year}{2020}\natexlab{b}.
\newblock \bibinfo{title}{Baba is Y'all: Collaborative Mixed-Initiative Level
  Design}.
\newblock
\newblock
\showeprint[arxiv]{2003.14294}~[cs.HC]


\bibitem[Doran and Parberry(2011)]%
        {Doran2011-questsMMORPGs}
\bibfield{author}{\bibinfo{person}{Jonathon Doran} {and} \bibinfo{person}{Ian
  Parberry}.} \bibinfo{year}{2011}\natexlab{}.
\newblock \showarticletitle{A Prototype Quest Generator Based on a Structural
  Analysis of Quests from Four MMORPGs}. In
  \bibinfo{booktitle}{\emph{Proceedings of the 2nd International Workshop on
  Procedural Content Generation in Games}} (Bordeaux, France)
  \emph{(\bibinfo{series}{PCGames '11})}. \bibinfo{publisher}{ACM},
  \bibinfo{address}{New York, NY, USA}, Article \bibinfo{articleno}{1},
  \bibinfo{numpages}{8}~pages.
\newblock
\showISBNx{9781450308724}
\urldef\tempurl%
\url{https://doi.org/10.1145/2000919.2000920}
\showDOI{\tempurl}


\bibitem[Dormans and Bakkes(2011)]%
        {dormans2011generating}
\bibfield{author}{\bibinfo{person}{Joris Dormans} {and} \bibinfo{person}{Sander
  Bakkes}.} \bibinfo{year}{2011}\natexlab{}.
\newblock \showarticletitle{Generating missions and spaces for adaptable play
  experiences}.
\newblock \bibinfo{journal}{\emph{IEEE Transactions on Computational
  Intelligence and AI in Games}} \bibinfo{volume}{3}, \bibinfo{number}{3}
  (\bibinfo{year}{2011}), \bibinfo{pages}{216--228}.
\newblock


\bibitem[Eladhari et~al\mbox{.}(2014)]%
        {Eladhari2014-storymakinggames}
\bibfield{author}{\bibinfo{person}{Mirjam~P Eladhari},
  \bibinfo{person}{Philip~L. Lopes}, {and} \bibinfo{person}{Georgios~N.
  Yannakakis}.} \bibinfo{year}{2014}\natexlab{}.
\newblock \showarticletitle{Interweaving Story Coherence and Player Creativity
  through Story-Making Games}. In \bibinfo{booktitle}{\emph{Interactive
  Storytelling : 7th International Conference on Interactive Digital
  Storytelling, ICIDS 2014, Singapore, Singapore, November 3-6, 2014,
  Proceedings}} \emph{(\bibinfo{series}{Lecture Notes in Computer Science},
  \bibinfo{number}{8832})}. \bibinfo{pages}{73--80}.
\newblock
\showISBNx{978-3-319-12336-3}
\urldef\tempurl%
\url{https://doi.org/10.1007/978-3-319-12337-0_7}
\showDOI{\tempurl}


\bibitem[Elson(2012)]%
        {elson-2012-dramabank}
\bibfield{author}{\bibinfo{person}{David Elson}.}
  \bibinfo{year}{2012}\natexlab{}.
\newblock \showarticletitle{{D}rama{B}ank: Annotating Agency in Narrative
  Discourse}. In \bibinfo{booktitle}{\emph{Proceedings of the Eighth
  International Conference on Language Resources and Evaluation ({LREC}'12)}}.
  \bibinfo{publisher}{European Language Resources Association (ELRA)},
  \bibinfo{address}{Istanbul, Turkey}, \bibinfo{pages}{2813--2819}.
\newblock
\urldef\tempurl%
\url{http://www.lrec-conf.org/proceedings/lrec2012/pdf/866_Paper.pdf}
\showURL{%
\tempurl}


\bibitem[Garbe et~al\mbox{.}(2019)]%
        {Garbe2019-StoryletsAssembler}
\bibfield{author}{\bibinfo{person}{Jacob Garbe}, \bibinfo{person}{Max
  Kreminski}, \bibinfo{person}{Ben Samuel}, \bibinfo{person}{Noah
  Wardrip-Fruin}, {and} \bibinfo{person}{Michael Mateas}.}
  \bibinfo{year}{2019}\natexlab{}.
\newblock \showarticletitle{StoryAssembler: An Engine for Generating Dynamic
  Choice-Driven Narratives}. In \bibinfo{booktitle}{\emph{Proceedings of the
  14th International Conference on the Foundations of Digital Games}} (San Luis
  Obispo, California, USA) \emph{(\bibinfo{series}{FDG '19})}.
  \bibinfo{publisher}{Association for Computing Machinery},
  \bibinfo{address}{New York, NY, USA}, Article \bibinfo{articleno}{24},
  \bibinfo{numpages}{10}~pages.
\newblock
\showISBNx{9781450372176}
\urldef\tempurl%
\url{https://doi.org/10.1145/3337722.3337732}
\showDOI{\tempurl}


\bibitem[García-Sánchez et~al\mbox{.}(2021)]%
        {tropesSimpsons}
\bibfield{author}{\bibinfo{person}{Pablo García-Sánchez},
  \bibinfo{person}{Antonio Velez-Estevez}, \bibinfo{person}{Juan
  Julián~Merelo}, {and} \bibinfo{person}{Manuel~Jesús Cobo}.}
  \bibinfo{year}{2021}\natexlab{}.
\newblock \showarticletitle{The Simpsons did it: Exploring the film trope space
  and its large scale structure}.
\newblock \bibinfo{journal}{\emph{PLOS ONE}} \bibinfo{volume}{16},
  \bibinfo{number}{3} (\bibinfo{date}{03} \bibinfo{year}{2021}),
  \bibinfo{pages}{1--28}.
\newblock


\bibitem[Gervas(2009)]%
        {Gervas2009-ComputationalStoryCreativity}
\bibfield{author}{\bibinfo{person}{Pablo Gervas}.}
  \bibinfo{year}{2009}\natexlab{}.
\newblock \showarticletitle{Computational Approaches to Storytelling and
  Creativity}.
\newblock \bibinfo{journal}{\emph{AI Magazine}} \bibinfo{volume}{30},
  \bibinfo{number}{3} (\bibinfo{date}{Jul.} \bibinfo{year}{2009}),
  \bibinfo{pages}{49}.
\newblock
\urldef\tempurl%
\url{https://doi.org/10.1609/aimag.v30i3.2250}
\showDOI{\tempurl}


\bibitem[Green et~al\mbox{.}(2021)]%
        {green2021-gamemechanicsAlignment}
\bibfield{author}{\bibinfo{person}{Michael~Cerny Green}, \bibinfo{person}{Ahmed
  Khalifa}, \bibinfo{person}{Philip Bontrager}, \bibinfo{person}{Rodrigo
  Canaan}, {and} \bibinfo{person}{Julian Togelius}.}
  \bibinfo{year}{2021}\natexlab{}.
\newblock \bibinfo{title}{Game Mechanic Alignment Theory and Discovery}.
\newblock
\newblock
\showeprint[arxiv]{2102.10247}~[cs.AI]


\bibitem[Greimas(1984)]%
        {Greimas84-structuralSemantics}
\bibfield{author}{\bibinfo{person}{Algirdas~Julien Greimas}.}
  \bibinfo{year}{1984}\natexlab{}.
\newblock \bibinfo{booktitle}{\emph{Structural Semantics: An Attempt at A
  Method}}.
\newblock \bibinfo{publisher}{University of Nebraska Press}.
\newblock


\bibitem[Hargood et~al\mbox{.}(2011)]%
        {Hargood2011-NarrativeCohesion}
\bibfield{author}{\bibinfo{person}{Charlie Hargood}, \bibinfo{person}{David
  Millard}, {and} \bibinfo{person}{Mark Weal}.}
  \bibinfo{year}{2011}\natexlab{}.
\newblock \showarticletitle{Measuring Narrative Cohesion: A Five Variables
  Approach}. In \bibinfo{booktitle}{\emph{Narrative and Hypertext Workshop at
  Hypertext 11 (05/06/11)}}.
\newblock
\urldef\tempurl%
\url{https://eprints.soton.ac.uk/272275/}
\showURL{%
\tempurl}


\bibitem[Harris(2016)]%
        {periodicTable}
\bibfield{author}{\bibinfo{person}{James Harris}.}
  \bibinfo{year}{2016}\natexlab{}.
\newblock \bibinfo{title}{Periodic Table of Storytelling}.
\newblock \bibinfo{howpublished}{\url{https://jamesharris.design/periodic/}}.
\newblock
\newblock
\shownote{Accessed: 2020-03-24}.


\bibitem[Hartsook et~al\mbox{.}(2011)]%
        {hartsook2011-storyWorlds}
\bibfield{author}{\bibinfo{person}{Ken Hartsook}, \bibinfo{person}{Alexander
  Zook}, \bibinfo{person}{Sauvik Das}, {and} \bibinfo{person}{Mark~O. Riedl}.}
  \bibinfo{year}{2011}\natexlab{}.
\newblock \showarticletitle{Toward supporting stories with procedurally
  generated game worlds}. In \bibinfo{booktitle}{\emph{2011 {IEEE} Conference
  on Computational Intelligence and Games, {CIG} 2011, Seoul, South Korea,
  August 31 - September 3, 2011}},
  \bibfield{editor}{\bibinfo{person}{Sung{-}Bae Cho}, \bibinfo{person}{Simon~M.
  Lucas}, {and} \bibinfo{person}{Philip Hingston}} (Eds.).
  \bibinfo{publisher}{{IEEE}}, \bibinfo{pages}{297--304}.
\newblock
\urldef\tempurl%
\url{https://doi.org/10.1109/CIG.2011.6032020}
\showDOI{\tempurl}


\bibitem[Horswill(2016)]%
        {Horswill2016-DearLeaderTrope}
\bibfield{author}{\bibinfo{person}{I. Horswill}.}
  \bibinfo{year}{2016}\natexlab{}.
\newblock \showarticletitle{Dear Leader’s Happy Story Time: A Party Game
  Based on Automated Story Generation}. In
  \bibinfo{booktitle}{\emph{Proceedings of the EXAG Workshop at AIIDE}}.
\newblock


\bibitem[Khalifa et~al\mbox{.}(2018)]%
        {Khalifa2018}
\bibfield{author}{\bibinfo{person}{Ahmed Khalifa}, \bibinfo{person}{Scott Lee},
  \bibinfo{person}{Andy Nealen}, {and} \bibinfo{person}{Julian Togelius}.}
  \bibinfo{year}{2018}\natexlab{}.
\newblock \showarticletitle{Talakat: Bullet hell generation through constrained
  map-elites}. In \bibinfo{booktitle}{\emph{Proceedings of The Genetic and
  Evolutionary Computation Conference}}. ACM.
\newblock


\bibitem[Kreminski et~al\mbox{.}(2019)]%
        {Kreminski2019-StorySifter}
\bibfield{author}{\bibinfo{person}{Max Kreminski}, \bibinfo{person}{Melanie
  Dickinson}, {and} \bibinfo{person}{Noah Wardrip-Fruin}.}
  \bibinfo{year}{2019}\natexlab{}.
\newblock \showarticletitle{Felt: A Simple Story Sifter}. In
  \bibinfo{booktitle}{\emph{Interactive Storytelling}},
  \bibfield{editor}{\bibinfo{person}{Rogelio~E. Cardona-Rivera},
  \bibinfo{person}{Anne Sullivan}, {and} \bibinfo{person}{R.~Michael Young}}
  (Eds.). \bibinfo{publisher}{Springer International Publishing},
  \bibinfo{address}{Cham}, \bibinfo{pages}{267--281}.
\newblock
\showISBNx{978-3-030-33894-7}


\bibitem[Kreminski and Wardrip-Fruin(2018)]%
        {Kreminski2018-SketchingStorylets}
\bibfield{author}{\bibinfo{person}{Max Kreminski} {and} \bibinfo{person}{Noah
  Wardrip-Fruin}.} \bibinfo{year}{2018}\natexlab{}.
\newblock \showarticletitle{Sketching a Map of the Storylets Design Space}. In
  \bibinfo{booktitle}{\emph{Interactive Storytelling}},
  \bibfield{editor}{\bibinfo{person}{Rebecca Rouse}, \bibinfo{person}{Hartmut
  Koenitz}, {and} \bibinfo{person}{Mads Haahr}} (Eds.).
  \bibinfo{publisher}{Springer International Publishing},
  \bibinfo{address}{Cham}, \bibinfo{pages}{160--164}.
\newblock
\showISBNx{978-3-030-04028-4}


\bibitem[Kybartas and Bidarra(2016)]%
        {kybartas2016survey}
\bibfield{author}{\bibinfo{person}{Quinn Kybartas} {and}
  \bibinfo{person}{Rafael Bidarra}.} \bibinfo{year}{2016}\natexlab{}.
\newblock \showarticletitle{A survey on story generation techniques for
  authoring computational narratives}.
\newblock \bibinfo{journal}{\emph{IEEE Transactions on Computational
  Intelligence and AI in Games}} \bibinfo{volume}{9}, \bibinfo{number}{3}
  (\bibinfo{year}{2016}), \bibinfo{pages}{239--253}.
\newblock


\bibitem[Lankoski(2013)]%
        {Lankoski2013-StoryConsistencyInteresting}
\bibfield{author}{\bibinfo{person}{Petri Lankoski}.}
  \bibinfo{year}{2013}\natexlab{}.
\newblock \showarticletitle{Models for Story Consistency and Interestingness in
  Single-Player RPGs}. In \bibinfo{booktitle}{\emph{Proceedings of
  International Conference on Making Sense of Converging Media}} (Tampere,
  Finland) \emph{(\bibinfo{series}{AcademicMindTrek '13})}.
  \bibinfo{publisher}{Association for Computing Machinery},
  \bibinfo{address}{New York, NY, USA}, \bibinfo{pages}{246–253}.
\newblock
\showISBNx{9781450319928}
\urldef\tempurl%
\url{https://doi.org/10.1145/2523429.2523480}
\showDOI{\tempurl}


\bibitem[{Levenshtein}(1966)]%
        {Levenshtein96-editDistance}
\bibfield{author}{\bibinfo{person}{V.~I. {Levenshtein}}.}
  \bibinfo{year}{1966}\natexlab{}.
\newblock \showarticletitle{{Binary Codes Capable of Correcting Deletions,
  Insertions and Reversals}}.
\newblock \bibinfo{journal}{\emph{Soviet Physics Doklady}}
  \bibinfo{volume}{10} (\bibinfo{date}{Feb.} \bibinfo{year}{1966}),
  \bibinfo{pages}{707}.
\newblock


\bibitem[Li and Riedl(2010)]%
        {Li2010-plannerPlotAdapt}
\bibfield{author}{\bibinfo{person}{Boyang Li} {and} \bibinfo{person}{Mark~O.
  Riedl}.} \bibinfo{year}{2010}\natexlab{}.
\newblock \showarticletitle{An offline planning approach to game plotline
  adaptation}. In \bibinfo{booktitle}{\emph{Proc. AIIDE'10}}.
\newblock


\bibitem[{Liapis} et~al\mbox{.}(2019)]%
        {Liapis2019-OrchestratingGames}
\bibfield{author}{\bibinfo{person}{A. {Liapis}}, \bibinfo{person}{G.~N.
  {Yannakakis}}, \bibinfo{person}{M.~J. {Nelson}}, \bibinfo{person}{M.
  {Preuss}}, {and} \bibinfo{person}{R. {Bidarra}}.}
  \bibinfo{year}{2019}\natexlab{}.
\newblock \showarticletitle{Orchestrating Game Generation}.
\newblock \bibinfo{journal}{\emph{IEEE Transactions on Games}}
  \bibinfo{volume}{11}, \bibinfo{number}{1} (\bibinfo{year}{2019}),
  \bibinfo{pages}{48--68}.
\newblock


\bibitem[Liapis et~al\mbox{.}(2014)]%
        {Liapis2014-gameCreativity}
\bibfield{author}{\bibinfo{person}{Antonios Liapis},
  \bibinfo{person}{Georgios~N. Yannakakis}, {and} \bibinfo{person}{Julian
  Togelius}.} \bibinfo{year}{2014}\natexlab{}.
\newblock \showarticletitle{Computational Game Creativity}. In
  \bibinfo{booktitle}{\emph{Proceedings of the fifth International Conference
  on Computational Creativity (ICCC 2014)}},
  \bibfield{editor}{\bibinfo{person}{Hannu Toivonen}, \bibinfo{person}{Simon
  Colton}, \bibinfo{person}{Michael Cook}, {and} \bibinfo{person}{Dan Ventura}}
  (Eds.).
\newblock


\bibitem[Mateas and Stern(2003)]%
        {mateas2003-facade}
\bibfield{author}{\bibinfo{person}{Michael Mateas} {and}
  \bibinfo{person}{Andrew Stern}.} \bibinfo{year}{2003}\natexlab{}.
\newblock \showarticletitle{Integrating plot, character and natural language
  processing in the interactive drama Fa{\c{c}}ade}. In
  \bibinfo{booktitle}{\emph{Proceedings of the 1st International Conference on
  Technologies for Interactive Digital Storytelling and Entertainment
  (TIDSE-03)}}, Vol.~\bibinfo{volume}{2}.
\newblock


\bibitem[Migkotzidis and Liapis(2021)]%
        {Panagiotis2021-susketch}
\bibfield{author}{\bibinfo{person}{Panagiotis Migkotzidis} {and}
  \bibinfo{person}{Antonios Liapis}.} \bibinfo{year}{2021}\natexlab{}.
\newblock \showarticletitle{SuSketch: Surrogate Models of Gameplay as a Design
  Assistant}.
\newblock \bibinfo{journal}{\emph{IEEE Transactions on Games}}
  (\bibinfo{year}{2021}), \bibinfo{pages}{1--1}.
\newblock
\urldef\tempurl%
\url{https://doi.org/10.1109/TG.2021.3068360}
\showDOI{\tempurl}


\bibitem[Mouret and Clune(2015)]%
        {Mouret2015-MAPElites}
\bibfield{author}{\bibinfo{person}{Jean-Baptiste Mouret} {and}
  \bibinfo{person}{Jeff Clune}.} \bibinfo{year}{2015}\natexlab{}.
\newblock \showarticletitle{Illuminating search spaces by mapping elites}.
\newblock \bibinfo{journal}{\emph{arXiv preprint arXiv:1504.04909}}
  (\bibinfo{year}{2015}).
\newblock


\bibitem[{Nintendo R\&D1} and {Intelligent Systems}(1985)]%
        {mario}
\bibfield{author}{\bibinfo{person}{{Nintendo R\&D1}} {and}
  \bibinfo{person}{{Intelligent Systems}}.} \bibinfo{year}{1985}\natexlab{}.
\newblock \bibinfo{title}{\emph{Super Mario Bros.}}
\newblock \bibinfo{howpublished}{Game [SNES]}.
\newblock
\newblock
\shownote{Nintendo, Kyoto, Japan. Last played August 2020.}.


\bibitem[{Nintendo R\&D4}(1991)]%
        {tloz:lttp}
\bibfield{author}{\bibinfo{person}{{Nintendo R\&D4}}.}
  \bibinfo{year}{1991}\natexlab{}.
\newblock \bibinfo{title}{\emph{The Legend of Zelda: A Link to the Past}}.
\newblock \bibinfo{howpublished}{Game [SNES]}.
\newblock
\newblock
\shownote{Nintendo, Kyoto, Japan. Last played December 2011.}.


\bibitem[{Nintendo R\&D4}(1998)]%
        {tloz:oot}
\bibfield{author}{\bibinfo{person}{{Nintendo R\&D4}}.}
  \bibinfo{year}{1998}\natexlab{}.
\newblock \bibinfo{title}{\emph{The Legend of Zelda: Ocarina of Time}}.
\newblock \bibinfo{howpublished}{Game [N64]}.
\newblock
\newblock
\shownote{Nintendo, Kyoto, Japan. Last played December 2011.}.


\bibitem[P{\'e}rez and Ortiz(2013)]%
        {Perez2013-AutomaticModelInterestingness}
\bibfield{author}{\bibinfo{person}{R.~Y. P{\'e}rez} {and}
  \bibinfo{person}{Otoniel Ortiz}.} \bibinfo{year}{2013}\natexlab{}.
\newblock \showarticletitle{A Model for Evaluating Interestingness in a
  Computer-Generated Plot}. In \bibinfo{booktitle}{\emph{Proceedings of the
  fourth International Conference on Computational Creativity (ICCC 2013)}}.
\newblock


\bibitem[Propp et~al\mbox{.}(1975)]%
        {propp1975-morphology}
\bibfield{author}{\bibinfo{person}{V. Propp}, \bibinfo{person}{L. Scott}, {and}
  \bibinfo{person}{L.A. Wagner}.} \bibinfo{year}{1975}\natexlab{}.
\newblock \bibinfo{booktitle}{\emph{Morphology of the Folktale: Second
  Edition}}.
\newblock \bibinfo{publisher}{University of Texas Press}.
\newblock
\showISBNx{9780292783911}
\showLCCN{68065567}
\urldef\tempurl%
\url{https://books.google.se/books?id=cyc7AQAAIAAJ}
\showURL{%
\tempurl}


\bibitem[Richmond and Schoentrup(2004)]%
        {tvtropes}
\bibfield{author}{\bibinfo{person}{Chris Richmond} {and} \bibinfo{person}{Drew
  Schoentrup}.} \bibinfo{year}{2004}\natexlab{}.
\newblock \bibinfo{title}{TV Tropes}.
\newblock
  \bibinfo{howpublished}{\url{https://tvtropes.org/pmwiki/pmwiki.php/Main/Tropes}}.
\newblock
\newblock
\shownote{Accessed: 2020-03-24}.


\bibitem[Riedl and Young(2006)]%
        {Riedl2006-StoryPlanningCreativity}
\bibfield{author}{\bibinfo{person}{Mark~O. Riedl} {and}
  \bibinfo{person}{R.~Michael Young}.} \bibinfo{year}{2006}\natexlab{}.
\newblock \showarticletitle{Story planning as exploratory creativity:
  Techniques for expanding the narrative search space}.
\newblock \bibinfo{journal}{\emph{New Generation Computing}}
  \bibinfo{volume}{24}, \bibinfo{number}{3} (\bibinfo{date}{01 Sep}
  \bibinfo{year}{2006}), \bibinfo{pages}{303--323}.
\newblock
\showISSN{1882-7055}
\urldef\tempurl%
\url{https://doi.org/10.1007/BF03037337}
\showDOI{\tempurl}


\bibitem[Rowe et~al\mbox{.}(2009)]%
        {Rowe2009-STORYEVAL}
\bibfield{author}{\bibinfo{person}{Jonathan Rowe}, \bibinfo{person}{Scott
  McQuiggan}, \bibinfo{person}{Jennifer Sabourin}, \bibinfo{person}{Derrick
  Marcey}, {and} \bibinfo{person}{James Lester}.}
  \bibinfo{year}{2009}\natexlab{}.
\newblock \showarticletitle{STORYEVAL: An empirical evaluation framework for
  narrative generation}.
\newblock \bibinfo{journal}{\emph{AAAI Spring Symposium - Technical Report}},
  \bibinfo{pages}{103--110}.
\newblock


\bibitem[Sarkar and Cooper(2021)]%
        {sarkar2021-dungeonPlatformer}
\bibfield{author}{\bibinfo{person}{Anurag Sarkar} {and} \bibinfo{person}{Seth
  Cooper}.} \bibinfo{year}{2021}\natexlab{}.
\newblock \showarticletitle{Dungeon and Platformer Level Blending and
  Generation using Conditional VAEs}. In \bibinfo{booktitle}{\emph{Proceedings
  of the IEEE Conference on Games (CoG)}}.
\newblock


\bibitem[Sharma et~al\mbox{.}(2007)]%
        {Sharna2007-PreferenceModelingStories}
\bibfield{author}{\bibinfo{person}{Manu Sharma}, \bibinfo{person}{Santiago
  Onta{\~{n}}{\'{o}}n}, \bibinfo{person}{Christina~R. Strong},
  \bibinfo{person}{Manish Mehta}, {and} \bibinfo{person}{Ashwin Ram}.}
  \bibinfo{year}{2007}\natexlab{}.
\newblock \showarticletitle{Towards Player Preference Modeling for Drama
  Management in Interactive Stories}. In \bibinfo{booktitle}{\emph{Proceedings
  of the Twentieth International Florida Artificial Intelligence Research
  Society Conference, May 7-9, 2007, Key West, Florida, {USA}}},
  \bibfield{editor}{\bibinfo{person}{David Wilson} {and} \bibinfo{person}{Geoff
  Sutcliffe}} (Eds.). \bibinfo{publisher}{{AAAI} Press},
  \bibinfo{pages}{571--576}.
\newblock
\urldef\tempurl%
\url{http://www.aaai.org/Library/FLAIRS/2007/flairs07-112.php}
\showURL{%
\tempurl}


\bibitem[Szilas(2002)]%
        {Szilas2003-structuralModelsIDtension}
\bibfield{author}{\bibinfo{person}{Nicolas Szilas}.}
  \bibinfo{year}{2002}\natexlab{}.
\newblock \showarticletitle{Structural models for Interactive Drama}. In
  \bibinfo{booktitle}{\emph{Proceedings of the 2nd International Conference on
  Computational Semiotics for Games and New Media}}.
\newblock
\urldef\tempurl%
\url{https://ci.nii.ac.jp/naid/10026187402/en/}
\showURL{%
\tempurl}


\bibitem[Szilas et~al\mbox{.}(2016)]%
        {Szilas2016-QualQuantInterestingness}
\bibfield{author}{\bibinfo{person}{Nicolas Szilas}, \bibinfo{person}{Sergio
  Estupi{\~{n}}{\'a}n}, {and} \bibinfo{person}{Urs Richle}.}
  \bibinfo{year}{2016}\natexlab{}.
\newblock \showarticletitle{Qualifying and Quantifying Interestingness in
  Dramatic Situations}. In \bibinfo{booktitle}{\emph{Interactive
  Storytelling}}, \bibfield{editor}{\bibinfo{person}{Frank Nack} {and}
  \bibinfo{person}{Andrew~S. Gordon}} (Eds.). \bibinfo{publisher}{Springer
  International Publishing}, \bibinfo{address}{Cham},
  \bibinfo{pages}{336--347}.
\newblock
\showISBNx{978-3-319-48279-8}


\bibitem[Tambwekar et~al\mbox{.}(2019)]%
        {tambwekar2019-controllableNeuralStory}
\bibfield{author}{\bibinfo{person}{Pradyumna Tambwekar},
  \bibinfo{person}{Murtaza Dhuliawala}, \bibinfo{person}{Lara~J. Martin},
  \bibinfo{person}{Animesh Mehta}, \bibinfo{person}{Brent Harrison}, {and}
  \bibinfo{person}{Mark~O. Riedl}.} \bibinfo{year}{2019}\natexlab{}.
\newblock \showarticletitle{Controllable Neural Story Plot Generation via
  Reward Shaping}. In \bibinfo{booktitle}{\emph{Proceedings of the
  Twenty-Eighth International Joint Conference on Artificial Intelligence,
  {IJCAI} 2019, Macao, China, August 10-16, 2019}},
  \bibfield{editor}{\bibinfo{person}{Sarit Kraus}} (Ed.).
  \bibinfo{publisher}{ijcai.org}, \bibinfo{pages}{5982--5988}.
\newblock
\urldef\tempurl%
\url{https://doi.org/10.24963/ijcai.2019/829}
\showDOI{\tempurl}


\bibitem[Thompson et~al\mbox{.}(2018)]%
        {Thompson2018-usingTropesNarrativeEvents}
\bibfield{author}{\bibinfo{person}{Matt Thompson}, \bibinfo{person}{Julian
  Padget}, {and} \bibinfo{person}{Steve Battle}.}
  \bibinfo{year}{2018}\natexlab{}.
\newblock \showarticletitle{Governing Narrative Events with Tropes as
  Institutional Norms}. In \bibinfo{booktitle}{\emph{Artificial Life and
  Intelligent Agents}}, \bibfield{editor}{\bibinfo{person}{Peter~R. Lewis},
  \bibinfo{person}{Christopher~J. Headleand}, \bibinfo{person}{Steve Battle},
  {and} \bibinfo{person}{Panagiotis~D. Ritsos}} (Eds.).
  \bibinfo{publisher}{Springer International Publishing},
  \bibinfo{address}{Cham}, \bibinfo{pages}{133--137}.
\newblock
\showISBNx{978-3-319-90418-4}


\bibitem[{Togelius} et~al\mbox{.}(2011)]%
        {Togelius2011}
\bibfield{author}{\bibinfo{person}{J. {Togelius}}, \bibinfo{person}{G.~N.
  {Yannakakis}}, \bibinfo{person}{K.~O. {Stanley}}, {and} \bibinfo{person}{C.
  {Browne}}.} \bibinfo{year}{2011}\natexlab{}.
\newblock \showarticletitle{Search-Based Procedural Content Generation: A
  Taxonomy and Survey}.
\newblock \bibinfo{journal}{\emph{IEEE Transactions on Computational
  Intelligence and AI in Games}} \bibinfo{volume}{3}, \bibinfo{number}{3}
  (\bibinfo{date}{Sept.} \bibinfo{year}{2011}), \bibinfo{pages}{172--186}.
\newblock
\showISSN{1943-068X}
\urldef\tempurl%
\url{https://doi.org/10.1109/TCIAIG.2011.2148116}
\showDOI{\tempurl}


\bibitem[Tolinsson et~al\mbox{.}(2020)]%
        {flodtol2020-WIPMakeSenseDungs}
\bibfield{author}{\bibinfo{person}{Simon Tolinsson}, \bibinfo{person}{Alexander
  Flodhag}, \bibinfo{person}{Alberto Alvarez}, {and} \bibinfo{person}{Jose
  Font}.} \bibinfo{year}{2020}\natexlab{}.
\newblock \showarticletitle{To Make Sense of Procedurally Generated Dungeons}.
  In \bibinfo{booktitle}{\emph{Extended Abstracts of the 2020 Annual Symposium
  on Computer-Human Interaction in Play}} (Virtual Event, Canada)
  \emph{(\bibinfo{series}{CHI PLAY '20})}. \bibinfo{publisher}{Association for
  Computing Machinery}, \bibinfo{address}{New York, NY, USA},
  \bibinfo{pages}{384–387}.
\newblock
\showISBNx{9781450375870}
\urldef\tempurl%
\url{https://doi.org/10.1145/3383668.3419890}
\showDOI{\tempurl}


\bibitem[Trenton et~al\mbox{.}(2010)]%
        {Trenton2010-questpatterns}
\bibfield{author}{\bibinfo{person}{Marcus Trenton}, \bibinfo{person}{Duane
  Szafron}, \bibinfo{person}{Josh Friesen}, {and} \bibinfo{person}{Curtis
  Onuczko}.} \bibinfo{year}{2010}\natexlab{}.
\newblock \showarticletitle{Quest Patterns for Story-Based Computer Games}. In
  \bibinfo{booktitle}{\emph{Proceedings of the Sixth AAAI Conference on
  Artificial Intelligence and Interactive Digital Entertainment}} (Stanford,
  California, USA) \emph{(\bibinfo{series}{AIIDE'10})}.
  \bibinfo{publisher}{AAAI Press}, \bibinfo{pages}{204–209}.
\newblock


\bibitem[van Stegeren and Myundefinedliwiec(2021)]%
        {vanstegeren2021-gpt2quests}
\bibfield{author}{\bibinfo{person}{Judith van Stegeren} {and}
  \bibinfo{person}{Jakub Myundefinedliwiec}.} \bibinfo{year}{2021}\natexlab{}.
\newblock \showarticletitle{Fine-Tuning GPT-2 on Annotated RPG Quests for NPC
  Dialogue Generation}. In \bibinfo{booktitle}{\emph{The 16th International
  Conference on the Foundations of Digital Games (FDG) 2021}} (Montreal, QC,
  Canada) \emph{(\bibinfo{series}{FDG'21})}. \bibinfo{publisher}{Association
  for Computing Machinery}, \bibinfo{address}{New York, NY, USA}, Article
  \bibinfo{articleno}{2}, \bibinfo{numpages}{8}~pages.
\newblock
\showISBNx{9781450384223}
\urldef\tempurl%
\url{https://doi.org/10.1145/3472538.3472595}
\showDOI{\tempurl}


\bibitem[Vozaru(2022)]%
        {vozaru_game_2022}
\bibfield{author}{\bibinfo{person}{Miruna Vozaru}.}
  \bibinfo{year}{2022}\natexlab{}.
\newblock \emph{\bibinfo{title}{The {Game} {Situation}: {An} object-based game
  analysis framework}}.
\newblock {PhD} {Thesis}. \bibinfo{school}{IT University of Copenhagen}.
\newblock
\newblock
\shownote{Published: PhD Dissertation, IT University of Copenhagen}.


\bibitem[Yannakakis et~al\mbox{.}(2014)]%
        {yannakakis2014micc}
\bibfield{author}{\bibinfo{person}{Georgios~N. Yannakakis},
  \bibinfo{person}{Antonios Liapis}, {and} \bibinfo{person}{Constantine
  Alexopoulos}.} \bibinfo{year}{2014}\natexlab{}.
\newblock \showarticletitle{Mixed-Initiative Co-Creativity}. In
  \bibinfo{booktitle}{\emph{Proceedings of the 9th Conference on the
  Foundations of Digital Games}}.
\newblock


\bibitem[Young et~al\mbox{.}(2013)]%
        {young2013-plansNarrGen}
\bibfield{author}{\bibinfo{person}{R.~Michael Young},
  \bibinfo{person}{Stephen~G. Ware}, \bibinfo{person}{Bradly~A. Cassell}, {and}
  \bibinfo{person}{Justus Robertson}.} \bibinfo{year}{2013}\natexlab{}.
\newblock \showarticletitle{Plans and planning in narrative generation: a
  review of plan-based approaches to the generation of story, discourse and
  interactivity in narratives}.
\newblock \bibinfo{journal}{\emph{Sprache und Datenverarbeitung, Special Issue
  on Formal and Computational Models of Narrative}} \bibinfo{volume}{37},
  \bibinfo{number}{1-2} (\bibinfo{year}{2013}), \bibinfo{pages}{41--64}.
\newblock


\bibitem[Yu et~al\mbox{.}(2020)]%
        {yu2020quest}
\bibfield{author}{\bibinfo{person}{Kristen Yu}, \bibinfo{person}{Nathan~R.
  Sturtevant}, {and} \bibinfo{person}{Matthew Guzdial}.}
  \bibinfo{year}{2020}\natexlab{}.
\newblock \showarticletitle{What is a Quest?}. In
  \bibinfo{booktitle}{\emph{Proceedings of the Intelligent Narrative
  Technolgies Workshop at AIIDE}}.
\newblock
\urldef\tempurl%
\url{http://www.cs.ualberta.ca/~nathanst/papers/yu2020quest.pdf}
\showURL{%
\tempurl}


\end{thebibliography}

\end{document}